\documentclass[letterpaper, 10 pt, conference]{ieeeconf}  
\IEEEoverridecommandlockouts                              
\usepackage{epsfig} 
\usepackage{times} 
\usepackage{hyperref}
\usepackage{mathtools}

\usepackage{amsmath} 
\usepackage{amssymb}  
\usepackage{mathrsfs}
\usepackage{algorithm,algorithmic}
\usepackage{ amssymb }
\usepackage{comment}
\title{\Large \bf
Orbit Characterization, Stabilization and Composition on 3D Underactuated Bipedal Walking via
 Hybrid Passive Linear Inverted Pendulum Model
}
    \author{Xiaobin Xiong and Aaron D. Ames
\thanks{*This work is supported by Amazon Fellowship in AI.}
\thanks{The authors are with the Department of Mechanical and Civil Engineering, California Institute of Technology, Pasadena, CA 91125
        {\tt\small xxiong@caltech.edu}, {\tt\small ames@caltech.edu}}%
 }
\begin{document}
\maketitle
\thispagestyle{empty}
\pagestyle{empty}

\begin{abstract}
A Hybrid passive Linear Inverted Pendulum (H-LIP) model is proposed for characterizing, stabilizing and composing periodic orbits for 3D underactuated bipedal walking. Specifically, Period-1 (P1) and Period-2 (P2) orbits are geometrically characterized in the state space of the H-LIP. Stepping controllers are designed for global stabilization of the orbits. Valid ranges of the gains and their optimality are derived. The optimal stepping controller is used to create and stabilize the walking of bipedal robots. An actuated Spring-loaded Inverted Pendulum (aSLIP) model and the underactuated robot Cassie are used for illustration. Both the aSLIP walking with P1 or P2 orbits and the Cassie walking with all 3D compositions of the P1 and P2 orbits can be smoothly generated and stabilized from a stepping-in-place motion. This approach provides a perspective and a methodology towards continuous gait generation and stabilization for 3D underactuated walking robots.  
 \end{abstract}
\section{INTRODUCTION}
Trajectory optimization has been widely used for planning periodic motion on 3D underactuated walking of bipedal robots. Advanced numerical techniques \cite{hereid20163d} \cite{posa2014direct} have been investigated for achieving fast performance. Model-based control methods \cite{ames2014rapidly} \cite{apgarfast} are developed for trajectory stabilization. However, trajectory optimization on the full dimensional model is still unlikely to be made fast enough to be performed online. The formulated nonlinear optimization is also subject to local minima. Moreover, stabilization on the optimized periodic motion is non-trivial, especially when the yielded \textit{hybrid zero dynamics} itself is not stable \cite{grizzle2014models}.

An alternative approach for motion planning for walking is through simplified conceptual models, e.g., the Linear Inverted Pendulum (LIP) \cite{pratt2012capturability} \cite{kajita2003biped} and the Spring-loaded Inverted Pendulum (SLIP) \cite{geyer2006compliant} \cite{rummel2010stable} models. The simplified models are viewed either as \textit{approximations} to the actual dynamics or as \textit{templates} for the systems to embed. 
For instance, the zero moment point (ZMP) approach \cite{kajita2003biped} can be viewed as embedding the template LIP dynamics into the robot's center of mass (COM) dynamics. The embedding approach has also appeared in the literature \cite{xiong2019exo} \cite{wensing2013high} \cite{Garofalo2012WalkingCO} \cite{Tyler} on the SLIP models. However, the direct embedding is problematic for underactuated walking due to the lack of full control authority on the COM.

When the simplified models are used as approximations to the full systems, they often serve for both planning and stabilization. For instance, the LIP model is used in \cite{pratt2012capturability} to plan the capture point for stabilization of the COM to desired states. An actuated SLIP model is proposed in \cite{xiong2018coupling} for planning the leg length trajectories for enabling and stabilizing walking in the sagittal plane of the robot Cassie. 

The stabilizations via simplified models are actually feedback planning on the desired trajectories to stabilize the COM dynamics. For instance, the SLIP model is used in \cite{raibert1986legged} for planning the footstep stabilization for the running velocity. Additional planning methods such as touch-down angle adjustment \cite{sharbafi2016vbla} and swing leg retraction \cite{seyfarth2003swing} have also been proposed in the literature. However, by and large, such feedback planning via simplified models is designed by heuristics and parameter tuning or searching is inevitable.

\begin{figure}[t]
      \centering
      \includegraphics[width= 2.7in]{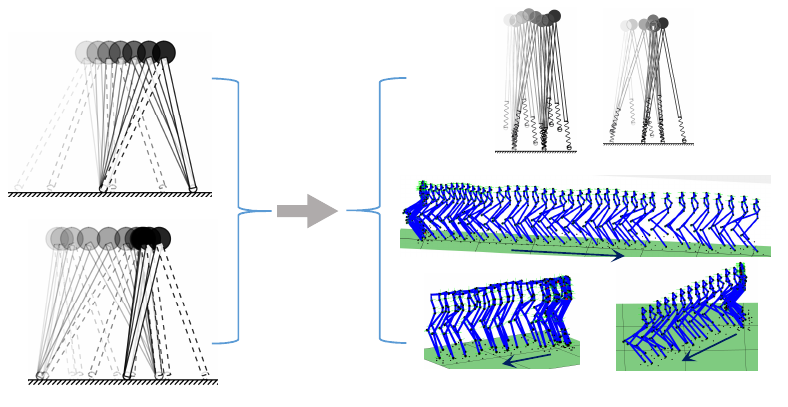}
      \caption{Overview of the stabilization from H-LIP model to aSLIP model and to the robot model of Cassie}
      \label{overview}
\end{figure}
\vspace{0.3cm}

\noindent {\bf Contributions.}
In this paper, we propose a Hybrid-Linear Inverted Pendulum (H-LIP) model for formal step planning for 3D underactuated walking without gain tuning. We first characterize its periodic orbits, including Period-1 (P1) and Period-2 (P2) orbits, and then synthesize velocity-based stepping controllers for global stabilization. We further prove the stabilization and identify the valid ranges of the feedback gains, including the optimal ones.

The H-LIP stepping stabilization can be used for its approximated complex walking systems. We first validate the H-LIP stepping stabilization on the walking of an actuated SLIP (aSLIP) \cite{xiong2018coupling}. The stepping stabilizes a stepping-in-place motion and generates different P1 and P2 orbits with a wide range of target velocities. Moreover, for 3D underactuated walking, the compositions of P1 and P2 orbits in the sagittal and coronal planes yield four types of walking behaviors. We apply the H-LIP stepping for constructing desired swing foot positions for the walking of the 3D robot model of Cassie \cite{xiong2018bipedal}. Walking gaits with all four compositions can be smoothly generated and stabilized from a stepping-in place motion to different target velocities. The proposed work thus provides a new methodology for characterizing and synthesizing continuous periodic walking for 3D underactuated bipedal walking via the H-LIP.

\section{Orbit Characterization and Stabilization for the Hybrid LIP Model}
The canonical Linear Inverted Pendulum (LIP) model is composed by a constant-height point mass attached on telescopic legs with actuated support pivots \cite{kajita2003biped}. 
It has been used as a target walking system to provide the desired COM dynamics of walking for the full robot to embed. The passive version of LIP (without actuation on the pivots) has also appeared in the literature \cite{pratt2012capturability} \cite{feng2016robust} \cite{faraji2014robust} for footstep planning to mitigate the overuse of ankle torques. \cite{xiong2018coupling} also suggests that the passive LIP dynamics can be used to approximate the lateral underactuated rolling dynamics of a full dimensional robot with passive feet. Here we formally present a hybrid version of the passive LIP (Hybrid-LIP) and study its periodic behaviors.  
\subsection{The Hybrid-LIP Model of Walking}
\begin{figure}[t]
      \centering
      \includegraphics[width= 3.3in]{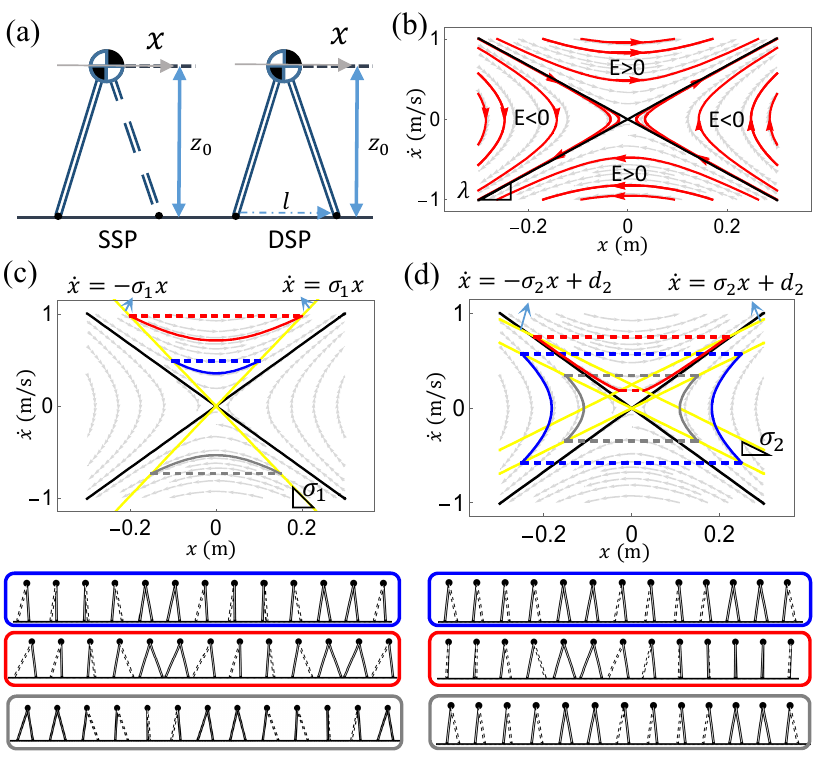}
      \caption{(a) Illustration of the two domain walking of the H-LIP model. (b) The phase portrait of the H-LIP in the SSP. (c) The P1 orbits of the H-LIP and the time-lapse figures of the orbits. (d) The P2 orbits of the H-LIP and the time-lapse figures of the orbits.}
      \label{LIPandOrbits}
\end{figure}

The Hybrid-LIP is the passive LIP model with two support legs. Based on the number of feet in contact with the ground, the walking of H-LIP is described by single support phase (SSP) and double support phase (DSP) (Fig. \ref{LIPandOrbits} (a)). We assume that in SSP the point mass dynamics is identical to that of the passive LIP, and that in DSP it has constant velocity. The dynamics can thus be written as,
\begin{align}
 \ddot{x} &= \lambda^2 x,   \tag{SSP}  \\
 \ddot{x} &= 0,   \tag{DSP}
\end{align}
where $\lambda = \sqrt{ \frac{g}{z_0}} $, and $z_0$ is the nominal height of the point mass. The transition from SSP to DSP, $\Delta_{\textrm{S} \rightarrow \textrm{D}}$, and the transition from DSP to SSP,  $\Delta_{\textrm{D} \rightarrow \textrm{S}}$, are assumed to be smooth, thus the impact maps are defined as:
\begin{eqnarray}
\label{ImpactS2D}
 \Delta_{\textrm{S} \rightarrow \textrm{D}} &:& \left \{\begin{matrix}
\dot{x}^{+} = \dot{x}^{-}  \\
x^{+} = x^{-}
\end{matrix}\right.,
\\
\label{ImpactD2S}
\Delta_{\textrm{D} \rightarrow \textrm{S}} &:& \left \{\begin{matrix*}[l]
\dot{x}^{+} = \dot{x}^{-}  \\
x^{+} = x^{-}   - l
\end{matrix*}\right.,
\end{eqnarray}
where $l$ is the step length from the stance foot position to the landing foot. The second map of the position is changing the support leg. The transitions are assumed to be time-based; in other words, the durations of the two domains, $\{T_\textrm{SSP}, T_\textrm{DSP}\}$, are fixed. The closed-form solution can be found:
\begin{eqnarray}
\label{LIPsol}
\textrm{SSP} &:& \left \{\begin{matrix*}[l]
x(t)= c_1 e^{\lambda t} + c_2 e^{- \lambda t}  \\
\dot{x}(t) = \lambda ( c_1 e^{\lambda t} - c_2 e^{- \lambda t})
\end{matrix*}\right.,
\\
\label{DSPsol}
\textrm{DSP} &:& \left \{\begin{matrix*}[l]
x(t) = x^{-}_{\textrm{SSP}} + \dot{x}^{-}_{\textrm{SSP}} t \\
\dot{x}(t) = \dot{x}^{-}_{\textrm{SSP}}
\end{matrix*}\right.,
\end{eqnarray}
where $ c_1 = \frac{1}{2} ( x^{+}_{\textrm{SSP}}+\frac{1}{\lambda}\dot{x}^{+}_{\textrm{SSP}}) $ and $ c_2 = \frac{1}{2} (  x^{+}_{\textrm{SSP}}-\frac{1}{\lambda}\dot{x}^{+}_{\textrm{SSP}}) $. Note that the H-LIP model can be in 3D. The dynamics in the each plane are identical and completely decoupled.

\textbf{Remark 1.1}
Compared to the passive LIP in \cite{pratt2012capturability} \cite{chevallereau2018self}, our H-LIP has the DSP and fixed domain durations. The addition of the DSP immediately rectifies the smooth impact assumption for compliant walking. The assumption on fixed domain durations will be self-evident in the later sections.

\subsection{Geometric Characterization of Periodic Orbits}
Periodic orbits are used to describe walking of the H-LIP. Here we study Period-1 (P1) and Period-2 (P2) orbits of the H-LIP. P1 orbits are the one-step orbits, i.e., states with each leg as the stance leg that evolve on the same orbits. P2 orbits are the two-step orbits. Fig. \ref{LIPandOrbits} (c)(d) show the examples of both orbits in the phase portrait of the H-LIP. The solid lines on the orbits represent the states in SSP, and the dashed straight lines on the orbits represent the jumps between SSP. The orbits in DSP is not explicitly represented for clarity.

The phase portrait of the H-LIP in SSP are divided into four regions by the asymptotes, i.e. the straight lines $\dot{x} = \pm \lambda x$ in Fig. \ref{LIPandOrbits} (b). The asymptotes characterize the states with zero \textit{orbital energy}, which is,
\begin{equation}
E(x, \dot{x}) = \dot{x}^2 - \lambda^2 x^2.
\end{equation}
For the H-LIP, the orbital energy is conserved in the SSP. The asymptotes also separate the state space into $E>0$ and $E<0$ regions. Since $\dot{x}^{+}_{\textrm{SSP}} =\dot{x}^{-}_{\textrm{SSP}}$, one can quickly verify that P1 orbits only exist in the $E>0$ regions, and that P2 orbits can exist in either $E>0$ or $E<0$ regions.

\subsubsection{Period-1 Orbits} For the ease of understanding, we first characterize the P1 orbits.

\textbf{Theorem 1.1} {\it For all the Period-1 orbits, the initial states $[x^{+}_{\textrm{SSP}}; \dot{x}^{+}_{\textrm{SSP}}]$ and the final states $[x^{-}_{\textrm{SSP}}; \dot{x}^{-}_{\textrm{SSP}}]$ are on the line $\dot{x} = -\sigma_1 x$ and the line $\dot{x} = \sigma_1 x$, respectively, where
\begin{equation}
\label{sigma1}
\sigma_{1} := \lambda \textrm{coth} \left(\frac{T_\textrm{SSP}}{2} \lambda\right),
\end{equation}
is called the \textit{orbital slope}. The lines $\dot{x} = \pm \sigma_1 x$ are called the \textit{orbital lines of characteristics}. Each state on the orbital line $\dot{x} = - \sigma_1 x$ is an initial state of the SSP of a unique Period-1 orbit with the step length being,
\begin{equation}
\label{stepLength1}
l_1 = x^{-}_{\textrm{SSP}} + \dot{x}^{-}_{\textrm{SSP}} T_{\textrm{DSP}} + \frac{\dot{x}^{-}_{\textrm{SSP}}}{\sigma_{1}}.
\end{equation}}

\textit{Proof.} Since $\dot{x}^{+}_{\textrm{SSP}} =\dot{x}^{-}_{\textrm{SSP}}$ and $E^{+}_{\textrm{SSP}} = E^{-}_{\textrm{SSP}}$, we have $x^{+}_{\textrm{SSP}} =-x^{-}_{\textrm{SSP}}$. Substituting these conditions into the Eq. \eqref{LIPsol} with simple algebra manipulation yields,
\begin{equation}
\frac{\dot{x}^{+}_{\textrm{SSP}} }{x^{+}_{\textrm{SSP}} } = -\lambda \textrm{coth}\left(\frac{T_\textrm{SSP}}{2} \lambda\right),
\end{equation}
which indicates that the initial states are on the line $\dot{x} = -\sigma_1 x$. The rest of the theorem follows immediately.  $ \blacksquare $

Fig. \ref{LIPandOrbits} (c) illustrates different P1 orbits of the system, each of which has a different net velocity. Without further illustration, one can find that there exist infinite number of P1 orbits, and they are all connected. Given a desired net velocity $v^{d}$, there exists a unique P1 orbit with the boundary velocity of the SSP being,
\begin{equation}\label{dxSSPdes1}
 \dot{x}^{-*}_{\textrm{SSP}} = \frac{v^{d}(T_{\textrm{DSP}} + T_{\textrm{SSP}})}{\frac{2}{\sigma_1}+T_{\textrm{DSP}}}.
\end{equation}
\subsubsection{Period-2 Orbits}
For P2 orbits, we index the support legs by its left leg $L$ and right leg $R$. By the constant velocity in DSP, $_{L}\dot{x}^{-}_{\textrm{SSP}} ={_{R}\dot{x}^{+}_{\textrm{SSP}}}$ and
$_{L}\dot{x}^{+}_{\textrm{SSP}} ={_{R}\dot{x}^{-}_{\textrm{SSP}}}$. The following theorem characterizes P2 orbits.

\textbf{Theorem 1.2} {\it For all the Period-2 orbits, the orbital lines of characteristics are $\dot{x} = \pm \sigma_{2} x+d_2
$, where the orbital slope $\sigma_2$ is,
\begin{equation}
\label{sigma2}
\sigma_{2} := \lambda {\textrm{tanh} \left(\frac{T_\textrm{SSP}}{2} \lambda \right)},
\end{equation}
and $d_2$ is a constant offset determined by the desired net velocity. Each state on the line $\dot{x} = - \sigma_{2} x+d_2$ represents a Period-2 orbit, with the step length,
\begin{equation}
\label{stepLength2}
l_2 =x^{-}_{\textrm{SSP}} +  T_{\textrm{DSP}}\dot{x}^{-}_{\textrm{SSP}}  + \frac{\dot{x}^{-}_{\textrm{SSP}}   - d_2}{\sigma_2}.
\end{equation}}

By the definition of orbital lines of characteristics, the immediate interpretation of the theorem is that: the initial states $[x^{+}_{\textrm{SSP}}; \dot{x}^{+}_{\textrm{SSP}}]$ lie on the line $\dot{x} = - \sigma_{2} x+d_2 $, and the final states $[x^{-}_{\textrm{SSP}}; \dot{x}^{-}_{\textrm{SSP}}]$ lie on the line $\dot{x} = \sigma_{2} x+d_2$. We first present the following corollary to prove this theorem.

\textbf{Corollary 1.3} In SSP, any initial states on the line of $\dot{x} = - \sigma_{2} x+d_2 $ flow to the line $\dot{x} = \sigma_{2} x+d_2 $ after $t = T_\textrm{SSP}$, with $\sigma_2$ defined in Eq. \eqref{sigma2} and $d_2$ being a constant.

\textit{Proof.} This is evident by applying the closed form solution from Eq. \eqref{LIPsol} with $t =T_{\textrm{SSP}}$. Given $\dot{x}^{+}_{\textrm{SSP}} =- \sigma_{2} x^{+}_{\textrm{SSP}} + d_2$, one can verify that $\dot{x}^{-}_{\textrm{SSP}} - \sigma_{2} x^{-}_{\textrm{SSP}} - d_2 \equiv 0$. $ \blacksquare $

\textit{Proof of Theorem 1.2.} We start by selecting a state on the line $\dot{x} = -\sigma_{2} x+d_2 $ as the initial state of the SSP with the left leg being the support leg of the H-LIP, i.e.  $_{L}\dot{x}^{+}_{\textrm{SSP}} = -\sigma_{2} {_{L}x^{+}_{\textrm{SSP}}}+d_2 $. By Corollary 1.3, the final state satisfies $_{L}\dot{x}^{-}_{\textrm{SSP}} = \sigma_{2} {_{L}x^{-}_{\textrm{SSP}}}+d_2 $. Applying the impact map in Eq. \eqref{ImpactS2D} yields $_{L}\dot{x}^{+}_{\textrm{DSP}} = {_{L}\dot{x}^{-}_{\textrm{SSP}}}, {_{L}x^{+}_{\textrm{DSP}}} = {_{L}x^{-}_{\textrm{SSP}}} T_{\textrm{DSP}} -  \frac{_{L}\dot{x}^{-}_{\textrm{SSP}}   - d}{\sigma_2}$. After the DSP in Eq. \eqref{DSPsol}, with the impact map in Eq. \eqref{ImpactD2S},  $_{R}\dot{x}^{+}_{\textrm{SSP}} = {_{L}\dot{x}^{-}_{\textrm{DSP}}}, {_{R}x^{+}_{\textrm{SSP}}} =  - \frac{ _{L}\dot{x}^{-}_{\textrm{SSP}}   - d}{\sigma_2} =- \frac{ _{R}\dot{x}^{+}_{\textrm{SSP}}   - d}{\sigma_2} $, which indicates that the initial state of SSP with right leg as the support is on the line $\dot{x} = -\sigma_{2} x+d_2 $ again. Taking another step, the system goes back to its original state when left leg becomes the support. Therefore, the initial state on the line creates a P2 orbit with the step length in Eq. \eqref{stepLength2}. $ \blacksquare $

Unlike the P1 orbits, for which only one orbit can be found to achieve certain desired velocity, there are infinite P2 orbits to achieve a certain desired velocity. The result is stated in the following proposition.

\textbf{Proposition 1.4} There exist infinite Period-2 orbits to achieve a desired net velocity $v_d$. The initial states of the SSP of all the periodic orbits for $v^d$ lie on the line $\dot{x} = -\sigma_{2} x+d_2 $ with $d_2$ being,
\begin{equation}
\label{d_2}
d_2 = \frac{\lambda^2 \textrm{sech} (\frac{T_{\textrm{SSP}}}{2}\lambda) (T_{\textrm{SSP}} + T_{\textrm{DSP}}) v_{d}}{\lambda^2 T_{\textrm{DSP}} + 2\sigma_{2}}.
\end{equation}
\textit{Proof.} Based on Theorem 1.2, selecting any initial state $[{_{L}x^{+}_{\textrm{SSP}}}, {_{L}\dot{x}^{+}_{\textrm{SSP}}}]$ on the line  $\dot{x} = -\sigma_{2} x+d_2$ yields a P2 orbit. The traveled distance over the P2 orbit is $2( {_{L}x^{-}_{\textrm{SSP}}} - {_{L}x^{+}_{\textrm{SSP}}}) + T_\textrm{DSP}( {_{L}\dot{x}^{+}_{\textrm{SSP}}} +  {_{L}\dot{x}^{-}_{\textrm{SSP}}})$. It equals to $2(T_{\textrm{SSP}} + T_{\textrm{DSP}}) v_{d}$. $ \blacksquare $


\textbf{Orbit Composition.} When the H-LIP is in 3D, the orbits can be selected individually for each plane. The composition of P1 and P2 orbits in each plane consequently yields four categories of walking orbits, i.e. P1-P2, P1-P1, P2-P1, and P2-P2 orbits. This will be further illustrated later in the 3D underactuated walking of Cassie.

\textbf{Equivalent Characterization.} It can also be shown that $\dot{x} = \pm \sigma_2 x +d_2$ can characterize P1 orbits, and that $\dot{x} =\pm \sigma_1 (x +d_1)$ can characterize P2 orbits subject to certain conditions. This part is omitted due to the space limit.
\subsection{Orbit Stabilization via Stepping}
In this part, we present the stabilization of the orbits of the H-LIP. Since the system is purely passive in each domain, the stabilization can only be done by manipulating the impact map, i.e. changing the step length $l$. We present the following two theorems for the stabilization.

\textbf{Theorem 2.1} {\it Given a desired Period-1 orbits with $\sigma_1$ in Eq. \eqref{sigma1}, $\dot{x}^{-*}_{\textrm{SSP}}$ in Eq. \eqref{dxSSPdes1} and $l_1$ in Eq. \eqref{stepLength1}, the following step length,
 \begin{equation}
\label{stepLengthcl1}
l_1^{cl} =l_1  + K(\dot{x}^{-}_{\textrm{SSP}} - \dot{x}^{-*}_{\textrm{SSP}}),
\end{equation}
can globally stabilize the H-LIP with the $K$ in,
\begin{equation}
\label{Krange}
0< K< \frac{2}{\lambda} \textrm{csch}(T_\textrm{SSP} \lambda),
\end{equation}
and the optimal gain,
\begin{equation}
\label{optimalGain}
K^{*}= \frac{1}{\lambda} \textrm{csch}(T_\textrm{SSP} \lambda),
\end{equation}
globally stabilizes the velocity by one step and the position by two steps.}

\textit{Proof.} The desired pre-impact state of SSP of the H-LIP is $[x^{-*}_{\textrm{SSP}} = \frac{\dot{x}^{-*}_{\textrm{SSP}}}{\sigma_1} ,  \dot{x}^{-*}_{\textrm{SSP}}]$. To prove this stabilization, we first show that the step to step velocity is contracting to the desired velocity $\dot{x}^{-*}_{\textrm{SSP}}$; in other words,
\begin{equation}
\label{constracting}
\left | {_{i+1}\dot{x}^{-}_{\textrm{SSP}}} - \dot{x}^{-*}_{\textrm{SSP}} \right | = c
\left | {_{i}\dot{x}^{-}_{\textrm{SSP}}} - \dot{x}^{-*}_{\textrm{SSP}} \right |,
\end{equation}
with $0<c<1$, where $i$ and $i+1$ index the current step and the next step, respectively. Suppose that we have an arbitrary preimpact state $[ {_{i}x^{-}_{\textrm{SSP}}},  {_{i}\dot{x}^{-}_{\textrm{SSP}}}]$. By the impact maps in Eq. \eqref{ImpactS2D} \eqref{ImpactD2S} with $l_1^{cl}$ and closed-form solutions in Eq. \eqref{LIPsol} \eqref{DSPsol}, the preimpact velocity of next step satisfies,
\begin{equation}
\label{velocityConv}
{_{i+1}\dot{x}^{-}_{\textrm{SSP}}}  - \dot{x}^{-*}_{\textrm{SSP}}= (1- K\lambda  \textrm{sinh}(T_\textrm{SSP} \lambda))( {_{i}\dot{x}^{-}_{\textrm{SSP}}} - \dot{x}^{-*}_{\textrm{SSP}}).
\end{equation}
Obviously, $K$ must be in the range in Eq. \eqref{Krange} so that the contracting mapping in Eq. \eqref{constracting} is satisfied. Otherwise, the velocity error stays the same or increases. Additionally, for the position,
\begin{equation}
\label{positionConv}
{_{i+1}x^{-}_{\textrm{SSP}}}  -  x^{-*}_{\textrm{SSP}} =( \frac{1}{\sigma_1} - K \textrm{cosh}(T_\textrm{SSP}  \lambda) ) ( {_{i}\dot{x}^{-}_{\textrm{SSP}}} -  \dot{x}^{-*}_{\textrm{SSP}}).
\end{equation}
Thus ${x^{-}_{\textrm{SSP}}} \rightarrow  x^{-*}_{\textrm{SSP}}$ as ${\dot{x}^{-}_{\textrm{SSP}}} \rightarrow  \dot{x}^{-*}_{\textrm{SSP}}$, which proves the stabilization of the position subsequently after the velocity.

Lastly, plugging the optimal gain of Eq. \eqref{optimalGain} into Eq. \eqref{velocityConv} and \eqref{positionConv} indicates that ${_{i+1}\dot{x}^{-}_{\textrm{SSP}}} =  \dot{x}^{-*}_{\textrm{SSP}}$ and ${_{i+2}x^{-}_{\textrm{SSP}}} =  x^{-*}_{\textrm{SSP}}$ for arbitrary states $[ {_{i}x^{-}_{\textrm{SSP}}},  {_{i}\dot{x}^{-}_{\textrm{SSP}}}]$, which means that, the velocity is stabilized by one step and the position is stabilized by two steps.
 $ \blacksquare $

 \textbf{Theorem 2.2} {\it Given a desired Period-2 orbit with $\sigma_2$ in Eq. \eqref{sigma2}, $l_2$ in Eq. \eqref{stepLength2} and desired boundary velocities ${_{L}\dot{x}^{-*}_{\textrm{SSP}}},{_{R}\dot{x}^{-*}_{\textrm{SSP}}}$, the following step length,
\begin{equation}
\label{stepLengthcl2}
l_2^{cl} = l_2 - K(\dot{x}^{-}_{\textrm{SSP}} - {_{L/R}\dot{x}^{-*}_{\textrm{SSP}}}),
\end{equation}
can stabilize the H-LIP globally with the $K$ in Eq. \eqref{Krange}. And the optimal gain in Eq. \eqref{optimalGain} stabilizes the velocity by one step and the position by two steps globally.}

The proof is similar to the previous one and thus is omitted. In our previous work \cite{xiong2018coupling}, the feedback stepping control law was used to stabilize the lateral balance of the underactuated walking of Cassie with zero lateral velocity, i.e. $d_2 = 0$. Here the stabilization law in Eq. \eqref{stepLengthcl2} is stated for the general case for all P2 orbits.
\begin{figure}[t]
      \centering
      \includegraphics[width= 3.3in]{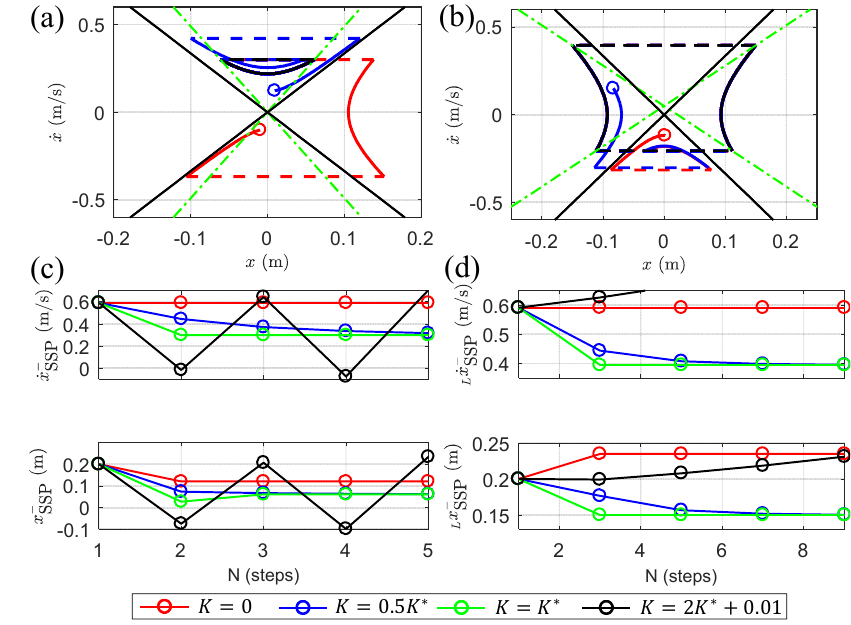}
      \caption{(a) Stabilization of the H-LIP to a desired P1 orbit from three random initial states. (b) Stabilization to a P2 orbit. (c) Comparison of the gain $K$ on $x^{-}_{\textrm{SSP}}, \dot{x}^{-}_{\textrm{SSP}}$ for stabilization to the P1 orbit from the same initial state. (d) Comparison of the gain $K$ for stabilization to the P2 orbit from the same initial state.}
      \label{LIPstb}
\end{figure}

Fig. \ref{LIPstb} (a) (b) illustrates the stabilization of this stepping with the optimal gain to a P1 and a P2 orbit from random initial states. Additionally, different gains are compared for the stabilization in Fig. \ref{LIPstb} (c) (d).

\textbf{Remark. 1.2} One may think that this stepping controller is a variant of the \textit{Raibert} style stepping controller \cite{raibert1986legged} \cite{rezazadeh2015spring}. However, the Raibert style controllers are oftentimes implemented as PID/PD terms. Heuristic tuning is required. Here the optimal gain eliminates the tuning process completely.
There are also additional interesting properties of the stabilization, which are omitted due to the space limit.


\section{aSLIP Walking via H-LIP Stepping}
The orbits of the H-LIP are all connected and can be stabilized. We posit the same characterization and stabilization can be readily applied to similar hybrid locomotion systems. In this section, we use an actuated Spring-loaded Inverted Pendulum (aSLIP) model \cite{xiong2018coupling} to illustrate this. The H-LIP stepping is a periodic behavior with constant durations, thus the aSLIP walking should start from a periodic behavior.

We first generate a periodic walking-in-place motion via trajectory optimization. The walking-in-place motion represents the origin of the H-LIP in its state space. Then we apply the walking stabilization based on H-LIP to stabilize the aSLIP walking smoothly from the origin to different orbits with different speeds. The periodic push-off and touch-down are preserved from the periodic actuation of the leg length from the walking-in-place motion. We demonstrate that both P1 and P2 orbits can be stabilized with negligible velocity errors. It is important to emphasize that the trajectory optimization is only performed once to generate a \textbf{single} gait. The rest of the walking orbits are naturally generated and stabilized via the H-LIP stepping.
\subsection{Hybrid Model of the aSLIP Walking}
The actuated Spring-loaded Inverted Pendulum (aSLIP) model in \cite{xiong2018coupling} was introduced to approximate the dynamics of the robot Cassie. Despite the mild complexity added comparing to canonical SLIP models, it well approximates the leg dynamics including the actuated parts and the underactuated compliant springs. Our previous work on hopping \cite{xiong2018bipedal} and walking \cite{xiong2018coupling} of the full robot Cassie have provided evidence for this. We posit that the aSLIP can also be applied for other pogo-stick style robots, such as MABLE \cite{SreenathPPG11}, ATRIAS \cite{rezazadeh2015spring}, ARL-Monopod \cite{ahmadi2006controlled}.

The main differences of the aSLIP compared to the canonical SLIP are two folds. First, the aSLIP model has a prismatic spring with nonlinear stiffness and damping, which are functions of the leg length $L$. Second, the actuation is enabled via the change of the leg length, i.e. $\ddot{L}$.

\begin{figure}[t]
      \centering
      \includegraphics[width= 2.8in]{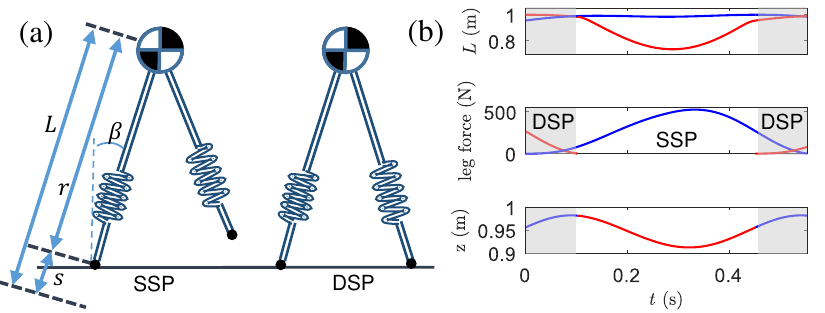}
      \caption{(a) Illustration of the two domain walking of the aSLIP model. (b) Trajectories of the leg length, force and mass height of the optimized walking-in-place gait.}
      \label{SLIP}
\end{figure}

The aSLIP walking is modeled as a hybrid dynamical system with two domains, i.e. SSP and DSP (Fig. \ref{SLIP} (a)). The transition from DSP to SSP happens when the ground reaction forces on one leg reach to 0; the transition from SSP to DSP happens when swing foot strikes the ground. The dynamics in each domain and the impact maps can be found in the Appendix.

\subsection{aSLIP Trajectory Optimization and Feedback Control}
We apply the same formulation of trajectory optimization via direct collocation in \cite{xiong2018coupling} to generate a periodic walking-in-place gait. The same periodic orbit will be also used for the robot Cassie in the next section.

In the trajectory optimization, we mainly enforce the domain constraints such as admissible ground reaction forces, swing foot positions, ranges of domain durations, zero walking speed and leg length ranges. The trivial swing leg dynamics is also included in optimization for periodic trajectories. The cost is on the virtually consumed energy, i.e. $\sum \ddot{L}^2$. Fig. \ref{SLIP} (b) shows one optimized gait of the aSLIP walking-in-place.

It is worth noting that the trajectory optimization can generate walking gaits with various step length or velocities. However, in this paper, we only optimize for one single gait. The purpose is to demonstrate that infinite gaits can be generated from a single gait via the H-LIP stepping.

\begin{figure*}[t]
      \centering
      \includegraphics[width= 6.6in]{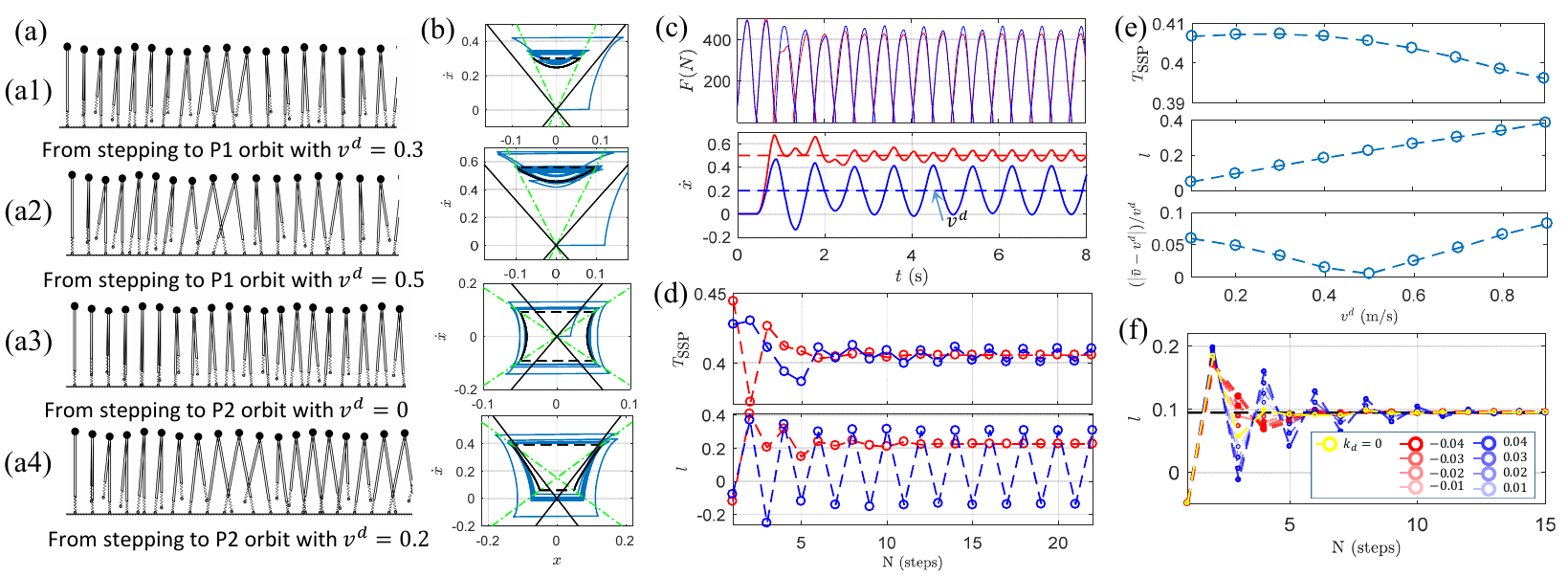}
      \caption{(a) The time-lapse figures of walking from stepping-in-place to various speeds. (b) The phase plots of the mass state $[x, \dot{x}]$ of the walking. The black orbits indicate the desired orbits of the H-LIP for the same velocities. (c) The leg forces $F$ and mass velocity $\dot{x}$ v.s. time of the (a2) (indicated by red) and (a4) (indicated by blue) walking. (d) The durations of the SSP $T_\textrm{SSP}$ and the step length $l$ of the (a2) and (a4) walking. (e) The converged $T_\textrm{SSP}$, $l$ and velocity error for P1 walking with speeds 0.1 to 0.9 m/s. (f) Comparison of adding derivative terms on stabilizing the walking with $v^d = 0.3$ m/s. }
      \label{SLIPstb}
\end{figure*}
The control of the aSLIP is realized by the zeroing the defined outputs. The optimized trajectories of the leg length $L(t)^d$ is used as the desired output trajectory for controlling the aSLIP walking. As we assume the $\ddot{L}_i$ are the inputs, trajectory tracking of the leg length can be realized via the linear feedback controller,
\begin{equation}
\ddot{L}_i = \ddot{L}^d_i - K_p ( L_i - L^{d}_i) - K_d ( \dot{L}_i - \dot{L}^{d}_i).
\end{equation}
Since the swing leg of the aSLIP in SSP has no dynamics, the step length can be directly set to the desired value. The following smoothing is used to transit the swing leg to the desired step length.
\begin{equation}
\label{stepLengthConstruction}
l(t) = (1 - c(t)) l_0 + c(t) l^d (t),
\end{equation}
where $l^d(t)$ is the desired step length, $c(t)$ is a smooth function from 0 to 1 with appropriate timing, and $l_0$ is the step length from previous step.

\subsection{Gait Stabilization and Generation via H-LIP stepping}
For the stepping-in-place, the desired velocity $v^d$ is 0. Both Eq. \eqref{stepLengthcl1} $l^{cl}_1$ and Eq. \eqref{stepLengthcl2} $l^{cl}_2$ from the H-LIP can be applied. In the state space $[x, \dot{x}]$ of the point mass, the stepping-in-place is represented by the origin. When a non-zero desired velocity $v^d$ is commanded, the H-LIP stepping stabilizes the aSLIP walking to the desired velocity. Fig. \ref{SLIPstb} (a) (b) shows the stabilization of the aSLIP walking from $0$ m/s to the desired non-zero velocities.

The stabilization via Eq. \eqref{stepLengthcl1} $l^{cl}_1$ automatically generates P1 orbits, and stabilization via Eq. \eqref{stepLengthcl2} $l^{cl}_2$ automatically generates P2 orbits. The generated P1 orbits are unique to a specific $v^d$ (Theorem 1.1 for H-LIP); while to a specific $v_d$ it can generate infinite number of P2 orbits, each of which corresponds to a specific boundary position $_{L/R} x^{-d}_{\textrm{SSP}}$ (Theorem 1.2). $_{L} x^{-d}_{\textrm{SSP}}$ is selected to be $- 0.05$m in thw walking in Fig. \ref{SLIPstb} (a3) (a4). Fig. \ref{SLIPstb} (c) (d) illustrate the convergence of the stabilization via H-LIP stepping in terms of the leg force, velocity, domain duration and step length.

Since the leg length repeats the same trajectory from the stepping-in-place optimization, the sum of the durations of the SSP and the DSP is guaranteed to be consistent across any orbits populated from the stepping-in-place orbit. It is expected that the leg internal behavior varies smoothly with respect to the variation of the forward velocity, so is the whole system behavior. Fig. \ref{SLIPstb} (e) shows the stabilization to P1 orbits with velocities from $0.1-0.9$ m/s. As expected, the converged $T_\textrm{SSP} \searrow$ smoothly and the leg length $l\nearrow$, as $v^d \nearrow$ smoothly. All converged velocity errors are within 10\%, indicating the success of gait generation and stabilization via H-LIP stepping.

\textbf{Comparison with a heuristic controller.} We also compare the H-LIP stepping stabilization with canonical Raibert style stepping controller \cite{poulakakis2009spring} \cite{raibert1986legged}, in the form of
$l^d = l_0 + K_p (v_{i} - v^d) +  K_d (v_{i} - v_{i-1})$, where $l_0$ is the nominal optimized step length, $K_p, K_d$ are the tunable proportional and derivative gains and $i, i-1$ index the current step and previous step.

Since the H-LIP contains a similar \textit{proportional} term, we only add the \textit{derivative} term to the H-LIP stepping (Eq. \eqref{stepLengthcl1}, \eqref{stepLengthcl2}). This can reduce the parameter space to be numerically explored. Fig. \ref{SLIPstb} (f) shows the comparison with different $K_d$ values. The optimal $K^* =0.1832$ from the H-LIP stepping; the $K_d$ is set within $[-0.04, 0.04]$. The step length is used as the criterion on the stabilization. As the results in the figure indicate, it is not evident if the \textit{derivative} term can improve the stabilization. In our opinion, the additional \textit{derivative} term is a \textit{placebo} that has existed in previous literature on bipedal walking, especially considering the facts: firstly, the H-LIP can already stabilize the system; secondly, inappropriate selection of $K_d$ can destabilize the system.

\textbf{Remark 1.3} Stabilization of the aSLIP walking is not a trivial task. The leg-axial oscillation needs to couple with the rotational oscillation about the support foot. Previously in \cite{xiong2018coupling}, we used a leg length scaling approach to update the desired leg length trajectories based on forward velocity errors, which worked well with phase-based output updates and nonzero velocity walking. The phase-based output update naturally creates a coupling between the rotational and leg-axial oscillations. The H-LIP stepping stabilization also creates an equivalent coupling via the state-dependent ($x, \dot{x}$) H-LIP stepping. Moreover, the H-LIP stepping propagates and stabilizes infinite orbits from a single stepping-in-place walking, which eliminates the tedious and extensive numerical optimization process.

\section{3D Underactuated Walking Stabilization, Generation and Composition via H-LIP Stepping}
The H-LIP stepping has shown success on the aSLIP walking generation and stabilization, despite that the aSLIP model is obviously not identical to the H-LIP model. The aSLIP is an approximated model of the robot Cassie \cite{xiong2018bipedal}, thus we apply the same H-LIP stepping on the walking of Cassie. Since the robot is in 3D, P1 and P2 orbits with their associated stabilizations can be selected for each plane separately, which further renders four kinds of compositions for walking.

Similarly, to apply the H-LIP stepping, the robot Cassie should start from a periodic behavior. The same periodic trajectory of leg length actuation of the aSLIP is applied to preserve the periodic push-off and touch-down behaviors. The H-LIP stepping provides the step sizes based on the selection of P1 and P2 orbits in each plane, which not only stabilizes the stepping-in-place motion but also generates different walking motion with different target velocities from the stepping-in-place walking.
\subsection{Robot Model and Hybrid Model of Walking}
\label{sec:Modeling}
The robot model of Cassie \cite{AG} has been detailed in previous paper \cite{xiong2018bipedal}. It is important to note that \cite{gong2018feedback} introduced a rigid model of Cassie for the ease of numerical gait generation, which leads to a 20 dof model with two holonomic constraints on each leg. In other words, the springs on the robot were assumed rigid in \cite{gong2018feedback}. Our robot model still includes all the \textbf{compliant springs}, which, in our viewpoint, are the main characteristics of the robot. The spring dynamics dominates the internal leg dynamics. In \cite{xiong2018bipedal} and \cite{xiong2018coupling}, we introduced the leg spring approximation for creating hopping and walking on Cassie, which also evidenced that the springs are critical.

As for the model of walking, we continue using the hybrid model of walking with two domains, i.e. SSP and DSP. The details can be found in \cite{xiong2018coupling}. The existence of the DSP naturally comes from the compliant springs in the legs. The two domain walking model also reflects back to the two domain walking of the H-LIP and the aSLIP.
\begin{figure*}[t]
      \centering
      \includegraphics[width = 6.95in]{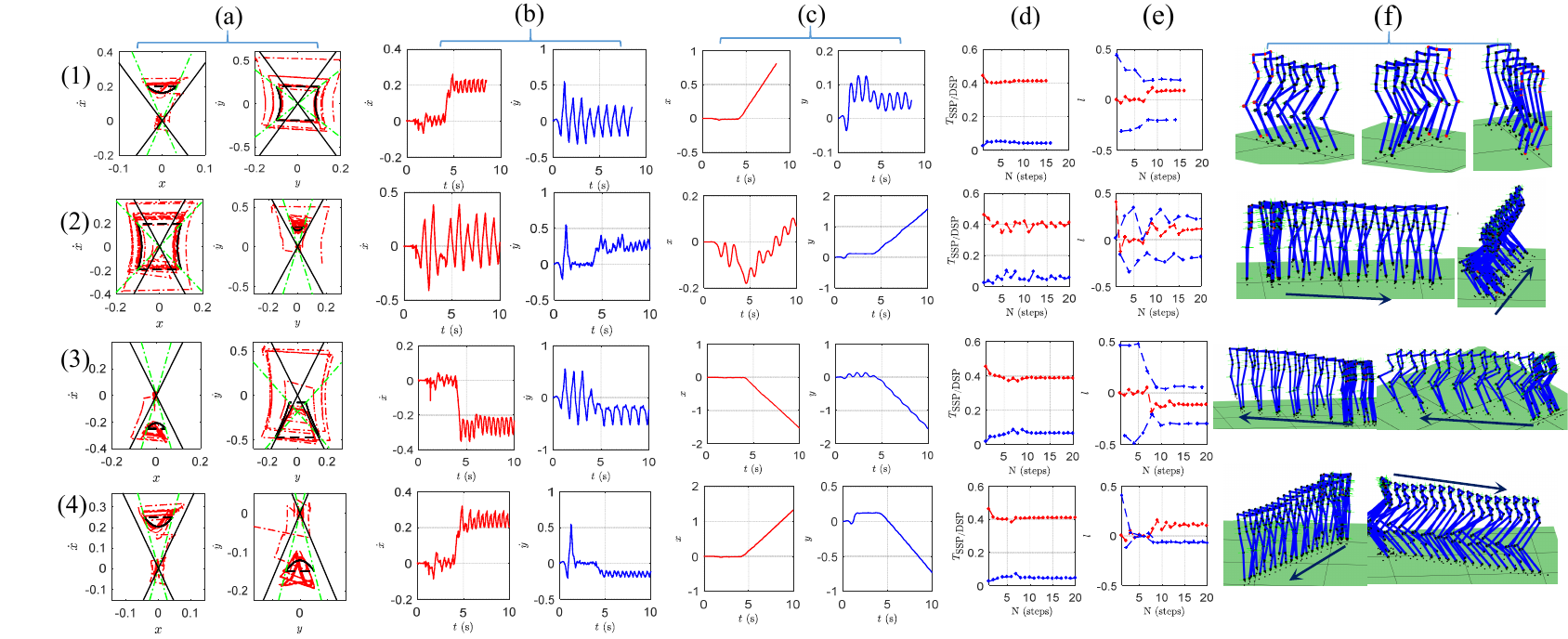}
      \caption{Simulation results of orbit composition and stabilization on Cassie via H-LIP stepping. (a) The phase plots of the robot in sagittal and lateral planes for stabilization of walking to (1: $\dot{x} \rightarrow 0.2, \dot{y} \rightarrow 0$), (2: $\dot{x} \rightarrow 0, \dot{y} \rightarrow 0.2$), (3: $\dot{x} \rightarrow -0.25, \dot{y} \rightarrow -0.25$), (4: $\dot{x} \rightarrow 0.25, \dot{y} \rightarrow -0.15$) from stepping-in-place. The black orbits are the ones of the H-LIP with same desired velocities. (b) The forward and lateral velocities $\dot{x}, \dot{y}$ v.s. time. (c) The forward and lateral positions $x, y$ v.s. time. (d) The convergence of the domain durations $T_\textrm{DSP}$ (blue), $T_\textrm{SSP}$ (red). (e) The convergence of the step length and step width in each plane. (f) The time-lapse figures of the walking motions. }
      \label{Cassieresults}
\end{figure*}
\subsection{Output Definition and Output Stabilization via CLF-QP}
We start from a stepping-in-place motion of Cassie. The motion is encoded via the definitions of the outputs. The leg length trajectory from the aSLIP is used on Cassie. Additional output definitions in SSP are similar to those described in \cite{xiong2018coupling}, such as the pelvis orientation, swing foot orientation, and the step width. The difference is that the swing leg angle output is replaced by the forward step length, which is the distance between two toes in the sagittal plane. The output definitions in DSP are identical to those in \cite{xiong2018coupling}.

The application of the H-LIP stepping is realized in SSP by the continuous construction of the desired step length in the sagittal plane and the desired step width in the coronal plane. The construction is similar to Eq. \eqref{stepLengthConstruction}. Feedback zeroing the outputs, i.e. trajectory tracking, is implementated by the control Lyapunov function based quadratic programs (CLF-QPs) \cite{xiong2018coupling} \cite{xiong2018bipedal}, which includes the torque limits and contact constraints in the QP.

\subsection{Gait Composition, Generation and Stabilization}
Now we generate and stabilize the walking of Cassie in 3D. We first compose the P1 and P2 orbits in the sagittal and coronal planes into 3D walking. The H-LIP stepping is then applied for stabilization and generation of walking based on the composition.

Both the P1 and P2 orbits can be realized in each plane. The composition yields four kinds of gaits in 3D. Let sP1-cP2 denote that a P1 orbit is realized in the sagittal plane and a P2 orbit in the coronal plane. Then the other three are sP1-cP1, sP2-cP2 and sP2-cP1. In general, all four gaits can be realized for a specific velocity of walking, without considering the kinematic constraints, which typically include that the step location shall be away from the stance location and that legs shall not cross. The avoidance of kinematic conflicts is out of the scope of this paper.

Fig. \ref{Cassieresults} illustrates the simulation results of the stabilization and generation of Cassie walking to different velocities with different orbit compositions. The stabilization starts from the stepping-in-place, then a nonzero velocity is commanded at about $t=3.5s$. For example, gait (1) is realized by a sP1-cP2 orbit with desired velocity ($\dot{x} \rightarrow 0.2, \dot{y} \rightarrow 0$), as shown in its phase plots. Gait (2) is realized by a cP1-sP2 orbit with desired velocity ($\dot{x} \rightarrow 0, \dot{y} \rightarrow 0.2$). Since the hybrid periodic walking behaviors of Cassie is similar to that of the H-LIP, the converged orbits are close to the ones of the H-LIP, as shown in Fig. \ref{Cassieresults}(a). The velocities and step sizes also converge to the desired ones quickly. Additional simulation videos can be found in \cite{Supplementary}.
%
%

\section{Conclusion}
In this paper, a Hybrid passive Linear Inverted Pendulum (H-LIP) model is proposed for orbit characterization, stabilization, and composition for 3D underactuated bipedal walking. We demonstrate that the closed-form stepping controllers with the optimal gains from the H-LIP can stabilize the stepping-in-place walking and smoothly generate the walking to different desired velocities for the aSLIP model and the robot model of Cassie. The characterized H-LIP orbits can compose into four kinds of walking orbits for 3D walking. Stabilization and generation of the composed orbits with different desired velocities can be realized simply via the H-LIP stepping with no gain tuning.

The theoretical guarantee is yet to be established on the stabilization from the H-LIP to complex hybrid models of walking. We posit that the hybrid nature of alternating support legs is the key to the success. Moreover, the orbit characterization and stabilization on the 3D walking via the H-LIP stepping also provide interesting perspectives as follows.
%

\textbf{Towards Bipedal Gait Categorization.} Periodic gaits have been well categorized for quadrupedal locomotion \cite{gan2018dynamic}. In bipedal locomotion, gaits are normally directly interpreted via individual periodic orbits. In our approach, we describe the bipedal walking via the orbits of the center of mass (COM) in the sagittal and coronal planes. Through the P1 and P2 orbits of the H-LIP, the combination of two types of orbits in 3D can provide a metric of categorizing the periodic orbits of bipedal locomotion.

\textbf{Towards Gait Stabilization.} The H-LIP stepping proves equivalent stabilization on the aSLIP and the full Cassie model. We are confident that this same approach can be applied to other pogo-stick robots. Our hypothesis is that the differences between these models are insignificant comparing to the hybrid periodic nature of the walking. Additionally, the height of the COM is preferred not to vary significantly during walking. We think that the H-LIP stepping is also possible to function on other bipedal or quadrupedal locomotion systems subject to mild modifications.

\textbf{Towards Continuous Gait Synthesis.} In the hybrid zero dynamics (HZD) \cite{grizzle2014models} approach for underactuated bipedal walking, stable orbits are constructed discretely, which consequently motivated the work of gait libraries \cite{da20162d}. In our approach, all the individual walking orbits are viewed as connected and can be continuously stabilized. All the walking of the aSLIP and Cassie come from a single stepping-in-place optimization of the aSLIP model. The stabilized walking motion is generated via the H-LIP stepping by the selection of the P1 or P2 orbits with the desired walking velocities.
In the future, we will emphasize our effort on establishing the theoretical guarantee on the stabilization from the simplified conceptual model to the full dimensional robot model, despite that this approach has been widely applied and accepted. We also would like to extend the H-LIP stepping to other types of walking robots.

%
%
\addtolength{\textheight}{-0.cm}

\section*{APPENDIX}
The continuous dynamics of the aSLIP are,
\begin{eqnarray}
\textrm{SSP}:\left\{\begin{matrix}
\ddot{r}_1 = \tfrac{F_1}{m} - g \text{cos}(\beta_1) + r \dot{\beta}_1^2
\\ \ddot{\beta}_1 =\tfrac{1}{r_1} ( -2\dot{\beta_1} \dot{r}_1 + g \text{sin}(\beta_1) )
\\  \ddot{s}_1 = \ddot{L}_1 - \ddot{r}_1
\end{matrix}\right., \nonumber \\
\textrm{DSP}:\left\{\begin{matrix}
\ddot{r}_1 = \frac{F_1 + F_2 \text{cos}(q_1 - q_2)}{m} - g \text{cos}(q_1) + r_1 \dot{q}_1^2  \\
 \ddot{q}_1 =\frac{1}{r_1} ( -2\dot{q}_1 \dot{r}_1 + g \text{sin}(q_1) - \frac{ F_2}{m} \text{sin}(q_1-q_2)  ) \\
  \ddot{r}_2 = \frac{F_2 + F_1 \text{cos}(q_1 - q_2)}{m} - g \text{cos}(q_2) + r_2 \dot{q}_2^2 \\
   \ddot{q}_2 =\frac{1}{r_2} ( -2\dot{q}_2 \dot{r}_2 + g \text{sin}(q_2) + \frac{ F_1}{m} \text{sin}(q_1-q_2)  )\\ \ddot{s}_1 = \ddot{L}_1 - \ddot{r}_1 \\
    \ddot{s}_2 = \ddot{L}_2 - \ddot{r}_2
\end{matrix}\right., \nonumber
\end{eqnarray}
where $\ddot{L}_1, \ddot{L}_2$ are assumed to be the inputs \cite{xiong2018coupling}. The spring forces $F_1, F_2$ come from the leg spring approximation.

\textit{Impact Map Assumption:}
We assume there is an impact event happening when the swing leg strikes the ground. Since the leg is massless, the velocity on the mass is assumed to be continuous. The velocity of the swing foot becomes zero at touchdown. Let 2 index the swing leg. $\dot{r}_2$ and $\dot{q}_2$ are discontinuous at impact. Additionally, the holonomic constraint on the swing leg enforces $\dot{L}^{+}_2 = \dot{s}^{+}_2 + \dot{r}^{+}_2$. It is assumed that the leg length velocity is continuous, i.e. $\dot{L}^{+}_2 = \dot{L}^{-}_2$. 
This matches with the intuition that, when the leg is rigidly controlled, the impact instantaneously acts on the compliant spring. The impact map is,
\begin{eqnarray}
\Delta_{\textrm{SSP} \rightarrow \textrm{DSP}}:\left\{\begin{matrix}
 \dot{q}^{+}_2 = \frac{1}{r_2}( \dot{q}_1 \dot{r}_1 \text{cos} (q_1 -q_2) + \dot{r}_1 \text{sin} (q_1 -q_2) )\\
 \dot{r}^{+}_2 = \dot{r}_1 \text{cos} (q_1 -q_2) -  \dot{q}_1 r_1  \text{sin} (q_1 -q_2)\\
 \dot{s}^{+}_2 = \dot{L}^{-}_2 - \dot{r}^{+}_2
 \end{matrix}\right.  \nonumber.
\end{eqnarray}
The transition map $\Delta_{\textrm{DSP} \rightarrow \textrm{SSP}}$ is smooth.
\bibliographystyle{IEEEtran}
\bibliography{Walking}
\end{document}